\documentclass[conference]{IEEEtran}
\IEEEoverridecommandlockouts
\usepackage{cite}
\usepackage{amsmath,amssymb,amsfonts}

\DeclareMathOperator*{\argmin}{arg\,min}

\usepackage{algorithmic}
\usepackage{graphicx}
\usepackage{textcomp}
\usepackage{xcolor}
\usepackage{microtype}
\usepackage{multirow}
\usepackage{booktabs, soul}
\usepackage{tabularx, color, soul}
\usepackage{comment}
\usepackage[colorlinks]{hyperref}       % hyperlinks
\usepackage{pifont}% http://ctan.org/pkg/pifont
\newcommand{\cmark}{\ding{51}}%
\newcommand{\xmark}{\ding{55}}%

\newcolumntype{C}{>{\centering\arraybackslash}X}
\def\BibTeX{{\rm B\kern-.05em{\sc i\kern-.025em b}\kern-.08em
    T\kern-.1667em\lower.7ex\hbox{E}\kern-.125emX}}
\usepackage{hyperref}
\usepackage[toc]{glossaries}
\makeglossaries

\newacronym{VVC}{VVC}{versatile video coding}
\newacronym{HEVC}{HEVC}{high efficiency video coding}
\newacronym{QoE}{QoE}{quality of experience}
\newacronym{HDR}{HDR}{high dynamic range}
\newacronym{HFR}{HFR}{high frame-rate}
\newacronym{UHD}{UHD}{ultra high definition}
\newacronym{JVET}{JVET}{joint video exploration team}
\newacronym{ITU-T}{ITU-T}{international telecommunication union}
\newacronym{MPEG}{MPEG}{moving picture expert group}
\newacronym{AI}{AI}{artificial intelligence}
\newacronym{ANN}{ANN}{artificial neural network}
\newacronym{LR}{LR}{low-resolution}
\newacronym{HR}{HR}{high-resolution}
\newacronym{MTL}{MTL}{multitask learning}
\newacronym{QE}{QE}{quality enhancement}
\newacronym{SR}{SR}{super-resolution}
\newacronym{QP}{QP}{quantization parameter}
\newacronym{VTM}{VTM-5}{VVC test model}
\newacronym{CTC}{CTC}{common test conditions}
\newacronym{BD}{BD}{Bjontegaard-Delta}
\newacronym{HD}{HD}{high-definition}
\newacronym{ML}{ML}{machine learning}
\newacronym{RB}{RB}{residual block}
\newacronym{EDSR}{EDSR}{enhanced deep super-resolution}
\newacronym{PSNR}{PSNR}{peak signal to noise ratio}
\newacronym{SSIM}{SSIM}{structural similarity}
\begin{document}

\title{Multitask Learning for VVC Quality Enhancement and Super-Resolution}

\author{\IEEEauthorblockN{Charles Bonnineau$^{\star \dagger \ddagger}$, Wassim Hamidouche$^{\star \ddagger}$, Jean-Fran\c cois Travers$^\dagger$, Naty Sidaty$^\dagger$ and Olivier Deforges$^\ddagger$}
\IEEEauthorblockA{$^\star$IRT b$<>$com, Cesson-Sevigne, France, \\
$^\dagger$TDF, Cesson-Sevigne, France, \\
$^\ddagger$Univ Rennes, INSA Rennes, CNRS, IETR - UMR 6164, Rennes, France}}

%\author{\IEEEauthorblockN{Charles Bonnineau$^{\star \dagger \ddagger}$}
%\IEEEauthorblockA{\textit{$^\star$IRT b$<>$com}\\
%35510 Cesson-Sevigne, France \\
%firstname.lastname@b-com.com}
%\and
%\IEEEauthorblockN{Jean-François Travers$^{\dagger}$, Naty Sidaty$^{\dagger}$}
%\IEEEauthorblockA{\textit{$^\dagger$TDF} \\
%35510 Cesson-Sevigne, France \\
%firstname.lastname@tdf.fr}
%\and
%\IEEEauthorblockN{Wassim Hamidouche$^{\ddagger \star}$, Olivier Deforges$^{\ddagger}$}
%\IEEEauthorblockA{\textit{$^\ddagger$IETR Lab UMR CNRS 6164, INSA Rennes}\\
%35708 Rennes, France \\
%firstname.lastname@insa-rennes.fr}
%}
\maketitle

\begin{abstract}

The latest video coding standard, called \gls*{VVC}, includes several novel and refined coding tools at different levels of the coding chain. These tools bring significant coding gains with respect to the previous standard, \gls*{HEVC}. However, the encoder may still introduce visible coding artifacts, mainly caused by coding decisions applied to adjust the bitrate to the available bandwidth. Hence, pre and post-processing techniques are generally added to the coding pipeline to improve the quality of the decoded video. These methods have recently shown outstanding results compared to traditional approaches, thanks to the recent advances in deep learning. Generally, multiple neural networks are trained independently to perform different tasks, thus omitting to benefit from the redundancy that exists between the models. In this paper, we investigate a learning-based solution as a post-processing step to enhance the decoded \gls*{VVC} video quality. Our method relies on multitask learning to perform both quality enhancement and super-resolution using a single shared network optimized for multiple degradation levels. The proposed solution enables a good performance in both mitigating coding artifacts and super-resolution with fewer network parameters compared to traditional specialized architectures.

\end{abstract}


\section{Introduction}
In recent years, the amount of video data has considerably increased due to the massive usage of video in our lives. The recent progress in camera lens and hardware technology has permitted the emergence of new video formats, improving the \gls*{QoE} of the end users. These video formats, including \gls*{HDR}, \gls*{HFR} and \gls*{UHD} with 4K and 8K resolutions, allow for a more realistic and immersive visual experience by increasing the amount of details about the scene. Therefore, highly efficient coding solutions are needed to deliver all these new video formats on the distribution networks. In response to this challenge, the \gls*{JVET}, established by the \gls*{ITU-T} and \gls*{MPEG}, has developed a new video coding standard, called \gls*{VVC}~\cite{xu2019recent}. This latter enables around 40\% of bit-rate reduction over \gls*{HEVC}~\cite{sullivan2012overview} for the same visual quality \cite{sidaty2019compression}.

In parallel, video processing deep learning-based solutions have been developed in order to reach these requirements and accelerate the deployment of these new services. For instance, the authors in \cite{bonnineau2020versatile} have proposed a backward-compatible solution for 8K and 4K signals broadcast using \gls*{VVC} and super-resolution. In \cite{wang2019attention}, a feed-forward network is used as an in-loop filter to enhance the quality of the \gls*{VVC} decoded frames. In practice, these models are trained independently, although some of the computed features can be useful for other tasks. Moreover, several models are generally proposed to optimize the network for different types of input data, e.g., levels of degradation, input channels. All these redundant parameters need to be stored or transmitted several times, thus being inappropriate for devices with limited energy and computing resources (mobile phones, TV chipsets, etc.). In this paper, we propose a multitask learning-based method that performs two tasks: super-resolution and quality enhancement of \gls*{VVC} intra-coded frames using a single network. We also use a multi-QPs training strategy based on fine-tuning and prior information inspired by \cite{wang2019attention}. Our approach enables a significant reduction of the number of parameters while maintaining a good performance compared to dedicated solutions. 

The rest of this paper is organized as follows. Section \ref{sec:related_work} provides a brief overview of deep learning-based post-processing and multitask learning approaches. The proposed solution is described in the Section \ref{sec:proposed_method}. Section \ref{sec:experiments} presents a performance evaluation through different experiments and provides an analysis of the results. Finally, Section \ref{sec:conclusion} concludes this paper.

\begin{figure*}[t]
\centering
\includegraphics[width=\linewidth]{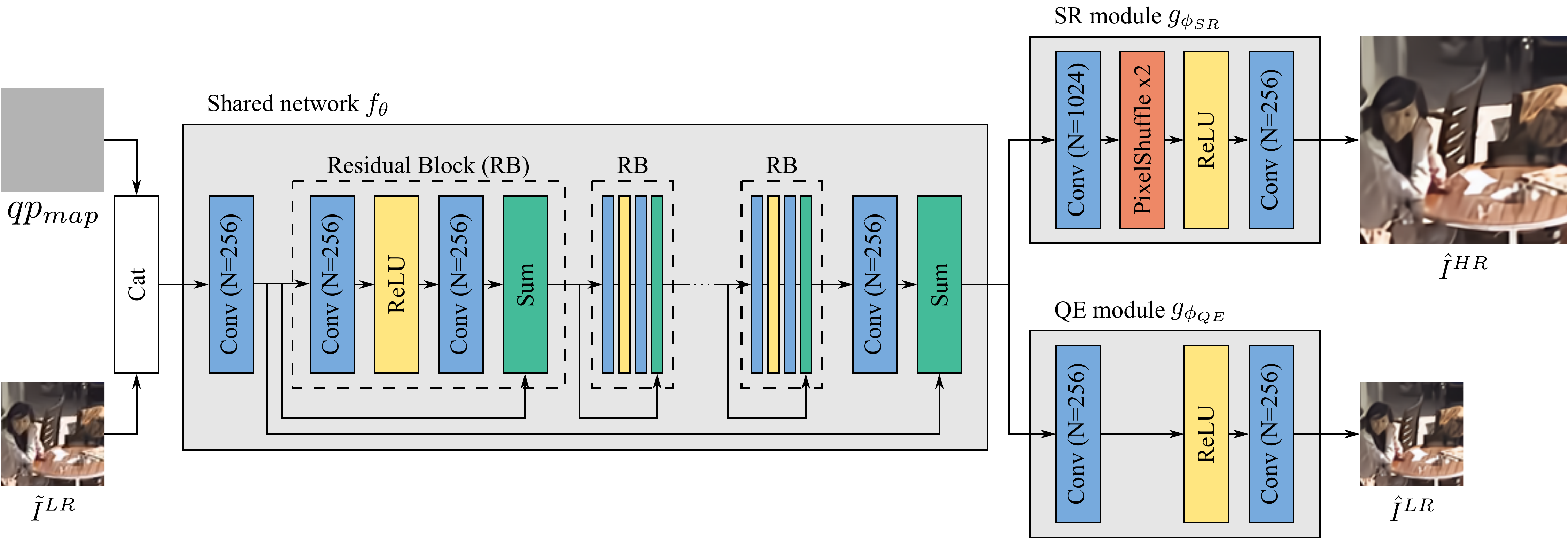}
\caption{The global pipeline of our multitask approach for super-resolution and quality enhancement.}
\label{fig:multitask}
\vspace{-0.2in}
\end{figure*}

\section{Related Work}
\label{sec:related_work}

With the recent progress in \gls*{ML}, a new type of learning-based super-resolution algorithms has emerged \cite{dong2014learning, shi2016real, lim2017enhanced}. These models are based on \gls*{ANN} and aim at learning the non-linear mapping that exists between \gls*{LR} and \gls*{HR} images. Those learning methods have outperformed the state-of-the-art up-scaling techniques, including hand-crafted interpolation filters \cite{keys1981cubic} and other learning approaches ranging from neighbor embedding \cite{chang2004super} to sparse coding \cite{yang2012coupled}. These architectures were also used without up-scaling for quality enhancement of reconstructed images to remove coding artifacts generated by lossy compression methods \cite{yu2016deep}. More advanced models were proposed, for instance, He {\it et al.}~\cite{he2018enhancing} considered the partition as a prior information to make the network focus on the block boundaries. In \cite{wang2019attention}, an architecture was developed with the use of an attention-based mechanism on pixels and channels to increase the quality enhancement efficiency.   

Recently, some deep models proposed the use of \gls*{MTL} \cite{caruana1997multitask} to perform multiple tasks using a single network. Therefore, the learned representations can be accessed by all the tasks in order to exploit redundant features and improve the performance. This concept has been essentially applied to high-level vision tasks \cite{liu2019end} where promising results were obtained. A first approach for super-resolution and quality enhancement has been proposed in \cite{zhang2018gated} using a gated fusion module. However, this method mainly focuses
on improving super-resolution applied to degraded images and does not maximize the information sharing between the tasks. 

\section{Proposed Method}
\label{sec:proposed_method}

%\subsection{Multi-degradation optimization}
%\begin{itemize}
%\item Finetuning
%\item Prior information
%\end{itemize}

%\begin{itemize}
%\item $lr^c$ : low-res compressed image
%\item $lr$ : low-res image
%\item $hr$ : high-res image
%\item $\hat{lr}$ : Enhanced image
%\item $\hat{hr}$ : Upscaled image
%\item $qp_{map}$ : QP map
%\item $l_{1}$ : L1 loss
%\item $L_{mtl}$ : Multitask loss
%\item $f^{SR}$ : SR-specific net
%\item $f^{Enh}$ : Enh-specific net
%\item $f^{s}$ : shared net
%\item $\theta_{SR}$ : SR-specific parameters
%\item $\theta_{Enh}$ : Enh-specific parameters
%\item $\theta_{s}$ : shared parameters
%\item $\hat{\theta}_{SR}$ : pre-trained SR-specific parameters
%\item $\hat{\theta}_{s}$ : pre-trained shared parameters
%\end{itemize}

Our approach aims to exploit the similarity between two tasks: \gls*{SR} and \gls*{QE}, with hard parameter sharing using a shared network $f_\theta$ and two task-specific modules $g_{\phi_{SR}}$ and $g_{\phi_{QE}}$. This method allows the model to benefit from the feature redundancy that exists between both tasks. Consequently, the number of parameters can be reduced while maintaining a good quality of reconstruction, compared to specialized architectures. 

To make the model capable of generalizing across several input \gls*{QP}, we use $qp_{map}$ \cite{wang2019attention} as prior information to the network. This prior input corresponds to a uniform normalized map computed as:

%Inspired by \cite{wang2019attention}, we use prior information about the \gls*{QP} to make the model capable of generalizing across several levels of degradation with a single multi-QPs training. This prior corresponds to a uniform normalized map computed as:

\begin{equation}
\label{eq:qp_map}
qp_{map}(i,j) = \frac{QP}{{QP}_{max}}, \quad i=1,...,W; \quad j=1,...,H,
\end{equation} 

with $(i,j)$ are the vertical and horizontal pixel coordinates. The value of ${QP}_{max}$ is equal to 63 in \gls*{VVC}. 

In the following, let $I^{LR}$ denote a low-resolution image of size $W \times H$ and $\tilde{I}^{LR}$ its reconstructed version that may include coding artifacts. We first extract the shared features $y$ from the input image $\tilde{I}^{LR}$ concatenated with its corresponding $qp_{map}$ using the shared network $f_\theta$ as follows:

\begin{equation}
    y = f_\theta (\tilde{I}^{LR} \oplus  qp_{map}),
\end{equation}

with $\oplus$ denoting the concatenation of $\tilde{I}^{LR}$ and $qp_{map}$.

The output images $\hat{I}^{HR}$ and $\hat{I}^{LR}$ are then estimated from the shared features $y$ using the task-specific modules $g_{\phi_{SR}}$ and $g_{\phi_{QE}}$ according to the following equations: 

\begin{equation}
    \hat{I}^{HR} = g_{\phi_{SR}} (y).
\end{equation}

\begin{equation}
    \hat{I}^{LR} = g_{\phi_{QE}} (y).
\end{equation}

We selected L1-loss~\cite{zhao2015loss} to compute the task-specific losses $\mathcal{L}_{SR}$ and $\mathcal{L}_{QE}$ between the estimated images $\hat{I}^{HR}$ and $\hat{I}^{LR}$, and the original images $I^{HR}$ and $I^{LR}$.

As our architecture is mainly inspired by \cite{lim2017enhanced}, we first pre-train the network to perform super-resolution on uncompressed images. The pre-trained parameters $\hat{\theta}$ and $\hat{\phi}_{SR}$ are obtained by solving the following optimization problem:

\begin{equation}
    (\hat{\theta};\hat{\phi}_{SR}) = \argmin_{(\theta;\phi_{SR})}\frac{1}{N}\sum_{n=1}^{N} \mathcal{L}_{SR}(g_{\phi_{SR}} (f_\theta (I^{LR}_n)),I^{HR}_n),
\end{equation}

with $I^{HR}_{n}$ the high-resolution training images, $I^{LR}_{n}$ the corresponding low-resolution versions, $N$ the number of training samples and $n=1,...,N$ the sample index

Finally, we optimize the overall multitask network by combining both task-specific losses in the multitask loss function $\mathcal{L}_{mtl}$ with a weighting parameter $\alpha$ as follows:

\begin{equation}
\label{eq:multitask_loss}
\mathcal{L}_{mtl}= \alpha \, \mathcal{L}_{SR}(\hat{I}^{HR}, I^{HR})+(1-\alpha) \, \mathcal{L}_{QE}(\hat{I}^{LR}, I^{LR}).
\end{equation}

The schematic in Fig.\ref{fig:multitask} illustrates the structure of the different components of our multitask network. The shared network $f_\theta$ mainly consists of $B$ \gls*{RB} with short and long skip connections. These operations allow the network to learn the identity function, improving the gradient flow from the deep to the shallow layers during the back-propagation step. It also leads to more sparse feature maps, and thus, better performance. For each convolutional layer, we use 256 filters of size $3 \times 3$. We introduce the non-linearity with the activation function ReLU between layers at different stages of the network. This structure is directly inspired by the \gls*{EDSR} network~\cite{lim2017enhanced}, which proposes state-of-the-art performance for super-resolution. We split the network at a very deep stage of the architecture to maximize the parameter sharing between tasks. For the super-resolution module $g_{\phi_{SR}}$, we use the Pixel-Shuffle upscaling layer \cite{shi2016real} at the end of the network. The same structure is used for the quality enhancement module $g_{\phi_{QE}}$, without the upscaling layer.

\section{Experimental results}
\label{sec:experiments}

\subsection{Training}
\label{sec:training}

%Methodology and reporting template for neural network coding tool testing JVET-M1006
%Our approach focus on the correction of VVC Intra-compression artefacts. 

%as suggested by \gls*{JVET} in \cite{jvet-m1006}

For the whole experiments, we train the networks with the DIV2K image dataset~\cite{agustsson2017ntire}. This later consists of 900 \gls*{HD} PNG pictures with a high diversity of spatial characteristics. To prevent network overfitting, we evaluate the performance on the Set5 image dataset~\cite{bevilacqua2012low}. The low-resolution images $I^{LR}$ are generated by a bicubic downscale applied on the high-resolution images $I^{HR}$. To generate the reconstructed versions $\tilde{I}^{LR}$ of the uncompressed images $I^{LR}$, we use the \gls*{VTM} in all-intra configuration with $QP\in\{22,27,32,37\}$ in order to simulate different levels of coding artifacts. We first convert the images from PNG to YUV4:2:0 format. Then, we collect the reconstructed images and convert them back to RGB. For training, we use  $64 \times 64$ patches extracted from the training set to reduce GPU memory usage. To test the performance of our network on video sequences, we also generate data from the ClassB and ClassA of the \gls*{JVET} \gls*{CTC}~\cite{jvet-m1010} using \gls*{VVC} all-intra, as described above. We also include two 8K videos, selected from the dataset given in \cite{jvet-q0791}. For the whole experiments, the quality is assessed on the luma component using \gls*{PSNR} and \gls*{SSIM}~\cite{wang2004image} image quality metrics computed between the estimated and original images. We also compute $\Delta$-\gls*{PSNR} and $\Delta$-\gls*{SSIM} that indicate the gain compared to the decoded images prior post-processing. For Super-Resolution, we use bicubic interpolation as anchor.

We train our model over 250 epochs, with a learning rate of $10^{-4}$ for from-scratch training and halve it for fine-tuning. For this latter, the pre-trained weights are obtained by training the network for super-resolution on uncompressed image pairs during 1000 epochs with a learning-rate of $10^{-4}$. We apply a learning rate decay with a gamma of 0.5 every 75 epochs to improve the convergence. We use a batch size of 8 and optimize the model with ADAM~\cite{kingma2014adam} by setting $\beta_1=0.9$, $\beta_2=0.999$ and $\epsilon=10^{-8}$. The parameter $\alpha$ in (\ref{eq:qp_map}) was tuned and fixed to 0.9 after a grid search on different values. All the experiments are performed on an NVIDIA Telsa V100 GPU using PyTorch.

%\begin{itemize}
%\item datasets
%\item training parameters
%\end{itemize}

\subsection{Multi-QPs optimization}
In the first experiment, we want to evaluate the ability of our multitask model to generalize across several QPs through an ablation study. The tested configurations include multi-QPs training, fine-tuning and $qp_{map}$, as described in Section \ref{sec:proposed_method}. Since less data are available for the training of the QP-specific networks than for a single multi-QPs network, we multiply the number of epochs by the number of tested QPs, i.e., 4, for these QP-specific configurations. We also adjust the learning rate decay to be applied every $4 \times 75$ epochs in this case. Thus, all the presented models are trained with the same number of parameter updates allowing a fair evaluation. The $qp_{map}$ is computed for each tested QP by (\ref{eq:qp_map}). We set the number of \gls*{RB} to $B=8$ for all the tested models, leading to around 13 million parameters per network.

%It allows us to evaluate our approach with high performance with a restrained number of parameters.

\begin{table}[t]
	\vspace{-1em}
	\scriptsize
	\begin{center}
		\caption{Ablation study of our model on Set5 for both tasks in terms of \gls*{PSNR} (dB) and $\Delta$-\gls*{PSNR} (dB).}
		\label{tab:multi_qp}
		\resizebox{0.98\columnwidth}{!}{
		\begin{tabularx}{\linewidth}{@{}lCCCCCCCC@{}}
			\midrule[0.3mm]
			Multi-QPs&\cmark&\cmark&\xmark&\xmark&\cmark&\cmark&\cmark&\cmark\\
			$qp_{map}$&\cmark&\cmark&\cmark&\cmark&\xmark&\xmark&\cmark&\cmark\\
			Fine-tuning&\cmark&\cmark&\cmark&\cmark&\cmark&\cmark&\xmark&\xmark\\
			Task  & SR & QE & SR & QE & SR & QE  & SR & QE \\
			\midrule[0.2mm]
			\multirow{2}{*}{QP22}&35.80 (+2.66)&43.07 (+0.43)&\textbf{35.85 (+2.71)}&\textbf{43.19 (+0.55)}&35.69 (+2.55)&42.92 (+0.28)&35.67 (+2.53)&42.98 (+0.34)\\
			\multirow{2}{*}{QP27}&34.17 (+1.91)&39.18 (+0.39)&\textbf{34.18 (+1.92)}&\textbf{39.24 (+0.45)}&34.16 (+1.90)&39.21 (+0.42)&34.09 (+1.83)&39.13 (+0.34)\\
			\multirow{2}{*}{QP32}&\textbf{32.16 (+1.22)}&35.62 (+0.40)&\textbf{32.16 (+1.22)}&\textbf{35.67 (+0.45)}&32.14 (+1.20)&35.63 (+0.41)&32.10 (+1.16)&35.58 (+0.36)\\
			\multirow{2}{*}{QP37}&29.85 (+0.70)&32.20 (+0.35)&\textbf{29.89 (+0.74)}&\textbf{32.27 (+0.42)}&29.78 (+0.63)&32.13 (+0.28)&29.81 (+0.66)&32.16 (+0.31)\\
			\midrule[0.2mm]
			\multirow{2}{*}{Average}&33.00 (+1.63)&37.52 (+0.39)&\textbf{33.02 (+1.65)}&\textbf{37.59 (+0.46)}&32.94 (+1.57)&37.47 (+0.34)&32.92 (+1.55)&37.46 (+0.33)\\
			\midrule[0.3mm]
		\end{tabularx}
		}
	\end{center}
	\vspace{-1em}
\end{table}

\begin{figure}[t]
\vspace{-0.1in}
\begin{minipage}[b]{0.49\linewidth}
  \centering
  \centerline{\includegraphics[width=1\linewidth]{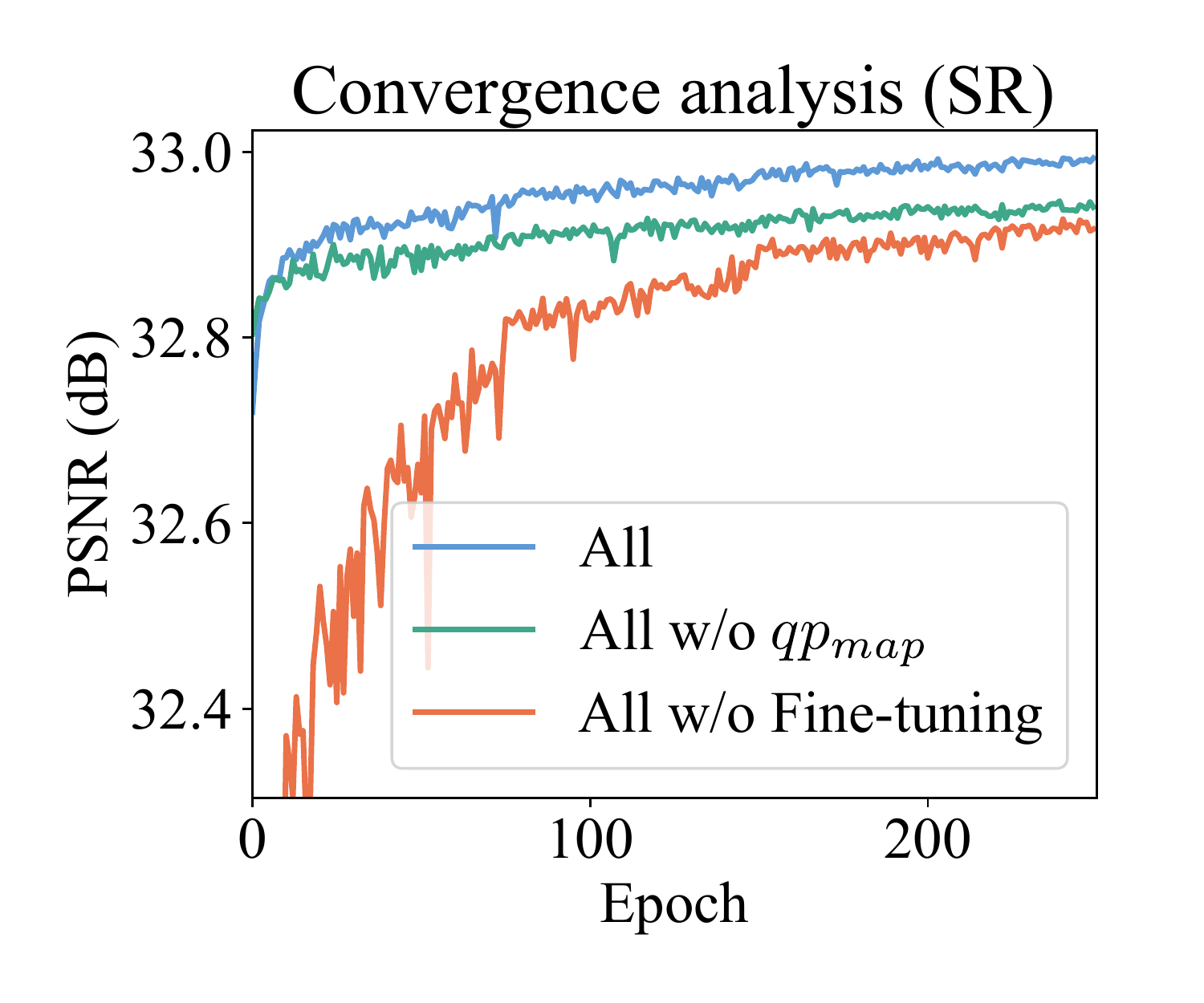}}
  \centerline{(a) Super resolution}\medskip
\vspace{-0.1in}
\end{minipage}
\hfill
\begin{minipage}[b]{0.49\linewidth}
  \centering
  \centerline{\includegraphics[width=1\linewidth]{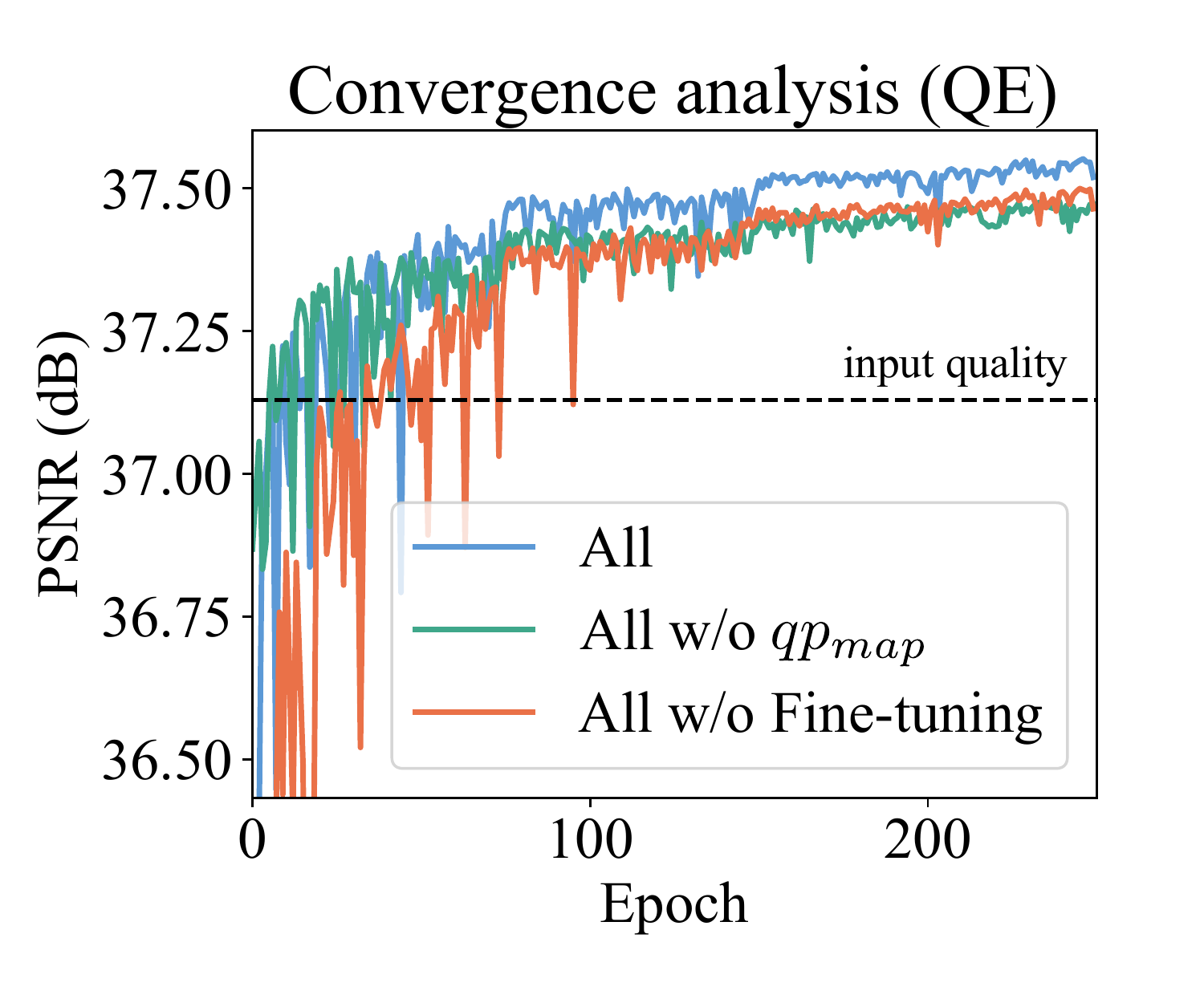}}
  \centerline{(b) Quality enhancement}\medskip
\vspace{-0.1in}
\end{minipage}
\hfill
\label{fig:convergence_analysis}
\caption{Convergence analysis of our multitask approach on Set5 regarding both tasks for different multi-QPs configurations.}
\vspace{-0.2in}
\end{figure}

Table \ref{tab:multi_qp} shows the performance of our model on Set5 dataset for different input \gls*{QP} in terms of \gls*{PSNR} (dB) and $\Delta$-\gls*{PSNR} (dB) for both super-resolution and quality enhancement. We perform an ablation study to evaluate the contribution of each component of our multi-\gls*{QP} model in the global performance of the network. We observe that a fine-tuning of the network pre-trained with uncompressed images leads to 0.08dB and 0.06dB of gain for SR and QE, respectively. We notice that even using parameters pre-trained for super-resolution, quality enhancement performs better as well. We also see that the $qp_{map}$ contributes to the performance of our multi-\gls*{QP} model by increasing the quality of reconstruction by 0.06dB for SR and 0.05dB for QE.  It can be noted that the models based on \gls*{QP}-specific training perform slighly better in terms of quality than our multi-\gls*{QP} model. However, training one network per \gls*{QP} requires four times more training time and parameters than a multi-QPs network to reach this level of performance.

Fig.~\ref{fig:convergence_analysis} visualizes the convergence of each multi-\gls*{QP} configuration by assessing the \gls*{PSNR} on the validation set at each training epoch for both tasks. We clearly notice that fine-tuning offers a more stable training with a faster convergence than from-scratch training for super-resolution. It is not surprising as the network starts to learn with weights that are already tuned for a related task. Although this configuration also leads to better results for quality enhancement, this observation is less pronounced in that case. Moreover, the training is globally less stable for this task. It can be explained by the fact that the loss related to super-resolution is more weighted in the proposed multitask loss $\mathcal{L}_{mtl}$. However, we notice that the use of $qp_{map}$ leads to a better convergence for both tasks.

\subsection{Multitask learning}

%bicubic	
%PSNR	SSIM
%33.1388095360616	0.912689349255733
%32.2581198299662	0.892163947136921
%30.9399423114615	0.861970081809741
%29.1523705405903	0.818859271713827
%31.3723105545199	0.871420662479056

%wo_enhancement	
%PSNR	SSIM
%42.6384062475808	0.987105298854035
%38.7937285516989	0.972616791135307
%35.2227637912655	0.948077530040615
%31.8508207238635	0.905219052881688
%37.1264298286022	0.953254668227911

\begin{table}[t]
	\vspace{-1em}
	\scriptsize
	\begin{center}
		\caption{Average performance (images, QPs) of the different Baselines in ($\Delta$-)\gls*{PSNR} (dB) and ($\Delta$-)\gls*{SSIM} computed on the Set5 dataset. The value of B corresponds to the number of \gls*{RB} used in the shared network $f_\theta$ for Baseline-B.}
		\label{tab:perf_multitask}
		\resizebox{0.98\columnwidth}{!}{
		\begin{tabularx}{\linewidth}{@{}lCCCCC@{}}
			\midrule[0.3mm]
			\multirow{2}{*}{\shortstack[l]{Method}}& \multirow{2}{*}{Baseline-B}&\multicolumn{2}{c}{Super-Resolution}& \multicolumn{2}{c}{Quality Enhancement}\\
			   &&	\gls*{PSNR} & \gls*{SSIM} & \gls*{PSNR} & \gls*{SSIM} \\
			\midrule[0.3mm]
			\multirow{4}{*}{Single-task SR}
			&\multirow{2}{*}{SR-4}&32.87 (+1.50)&0.8909 (+0.0195)& \multirow{2}{*}{|} & \multirow{2}{*}{|}\\
			&\multirow{2}{*}{SR-8}&\textbf{33.00 (+1.63)}&\textbf{0.8925 (+0.0211)}& \multirow{2}{*}{|} & \multirow{2}{*}{|} \\
			\midrule[0.2mm]
			\multirow{4}{*}{Single-task QE}
			&\multirow{2}{*}{QE-4}& \multirow{2}{*}{|} & \multirow{2}{*}{|} & 37.43 (+0.30)&0.9552 (+0.0020) \\
			&\multirow{2}{*}{QE-8}& \multirow{2}{*}{|} & \multirow{2}{*}{|} & 37.50 (+0.36)& 0.9557 (+0.0025) \\
			\midrule[0.2mm]
			\multirow{2}{*}{Sequential}&\multirow{2}{*}{QE-4;SR-4}&32.88 (+1.51)&0.8912 (+0.0198)&37.43 (+0.30)&0.9552 (+0.0020)\\
			\midrule[0.2mm]
			\multirow{2}{*}{Multitask}&\multirow{2}{*}{MTL-8}&\textbf{33.00 (+1.63)} & 0.8924 (+0.0210) & \textbf{37.52 (+0.39)} & \textbf{0.9558 (+0.0026)}\\
			\midrule[0.3mm]
		\end{tabularx}
	}
	\end{center}
	\vspace{-2.5em}
\end{table}

In this experiment, we demonstrate the effectiveness of our multitask approach compared to specialized networks. Single-task architectures derive from our multitask solution with $\alpha$ set to 0 and 1 in the multitask loss $\mathcal{L}_{mtl}$, defined in (\ref{eq:multitask_loss}), for quality enhancement and super-resolution, respectively. We also include a sequential configuration based on these two single-task networks. For an input image $\tilde{I}^{LR}$, the sequential configuration can be expressed as:

\begin{equation}
\hat{I}^{HR} = g_{\hat{\phi}_{SR}} (f_{\hat{\theta}} (g_{\phi_{QE}} (f_{\hat{\theta}} (\tilde{I}^{LR} \oplus qp_{map})))).
\end{equation}

In that case, quality enhancement is applied to the input image before passing through the super-resolution specialized network. For this experiment, we set the number of \gls*{RB} for each specialized network in both sequential and single-task configurations to $B=4$, leading to approximately 7.7 million parameters per network. We also include single-task models with $B=8$ to match the performance with the multitask configuration. All the models are trained using fine-tuning and $qp_{map}$. Table \ref{tab:perf_multitask} gives the performance in terms of \gls*{PSNR}, \gls*{SSIM}, $\Delta$-\gls*{PSNR} and $\Delta$-\gls*{SSIM} for both tasks.

Firstly, we can notice that the sequential configuration does not perform significantly better than the single-task by considering the same total number of parameters. Moreover, super-resolution with $B=8$ enables better performance than the sequential model regarding this task. For our multitask approach, we can notice that the performance of single-task is reached with half parameters. In addition, the multitask performs better in terms of quality than single-task considering the same total number of parameters. It can be explained by the fact that a large number of features are computed twice between the specialized architectures. Thus, a deeper multitask model allows new representations to be learned, increasing the quality of reconstruction for both tasks.

\begin{figure}[t]
\centering
\includegraphics[width=\linewidth]{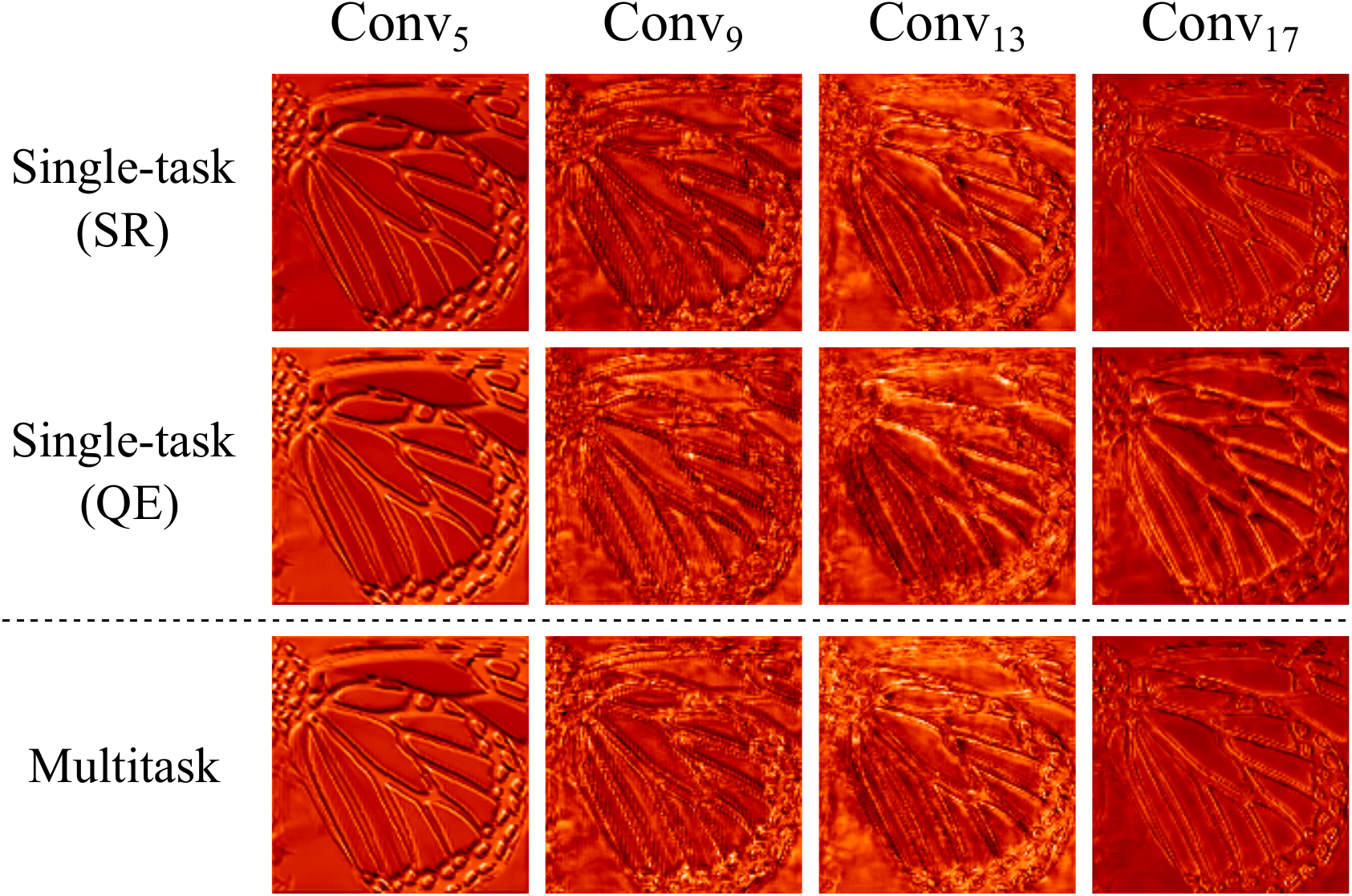}
\caption{Average feature maps at different stage of the baselines QE-8, SR-8 and MTL-8 for the input image \textit{butterfly.png} encoded with \gls*{VTM} AI (qp27). Conv\textsubscript{i} corresponds to the i-th convolutional layer of the network.}
\label{fig:feature_analysis}
\vspace{-0.2in}
\end{figure}

In Fig.\ref{fig:feature_analysis}, we display the average feature maps for different convolutional layers of the single-task and multitask architectures. As shown in this figure, the features are globally similar and become more complex and specialized in the deeper layers. For Conv\textsubscript{17}, the average feature map of multitask is more similar to super-resolution than quality enhancement, mostly because $\alpha=0.9$ in the multitask loss $\mathcal{L}_{mtl}$ of the proposed model. This demonstrates that a high correlation exists between the presented single-task models which can be exploited in the multitask architecture. 

\subsection{Coding performance}

In the last experiment, we investigated the performance of our multitask model applied as post-processing for video delivery against single-task networks. The input signal is first downscaled and encoded. Then, both post-processing tasks are performed on the decoded signal outside the coding loop, as presented in \cite{bonnineau2020versatile} for super-resolution. The bit-rate is assessed on the low-resolution signal. For this experiment, we consider the same total number of parameters for both tested configurations, i.e., $B=8$ for our multitask network and $B=4$ for each single-task network. We use the \gls*{BD}-Rate method described in \cite{vceg_m33} to evaluate our approach. Table \ref{tab:perf_multitask_bd} presents the results for both super-resolution and quality enhancement. 

We can notice that, in average, our multitask model allows 2.8\%/2.1\% and 2.3\%/1.1\% of bit-rate savings over specialized networks for the same objective quality, using \gls*{PSNR} and \gls*{SSIM}, regarding super-resolution and quality enhancement, respectively. We can also notice that these gains are higher for the sequences where our method performs well against naive anchors. These video sequences including { \it BQTerrace} and { \it SubwayTree} contain more spatial information and need more powerful models to be accurately reconstructed. 

\begin{table}[t]
	\vspace{-1em}
	\scriptsize
	\begin{center}
		\caption{BD-rate (\%) of our approach computed over single-task regarding \gls*{PSNR} and \gls*{SSIM} for different resolution classes. The values in bracket indicate the gain compared to naive anchors, i.e., bicubic upscale and input quality.}
		\label{tab:perf_multitask_bd}
		\resizebox{0.98\columnwidth}{!}{
		\begin{tabularx}{\linewidth}{@{}llCCCC@{}}
			\midrule[0.3mm]
			\multirow{2}{*}{Dataset}& \multirow{2}{*}{Sequence}&\multicolumn{2}{c}{Super-Resolution}& \multicolumn{2}{c}{Quality Enhancement}\\
			   &&	(\gls*{PSNR}) & (\gls*{SSIM}) & (\gls*{PSNR}) & (\gls*{SSIM}) \\
			\midrule[0.3mm]
			\multirow{4}{*}{\shortstack[l]{8K \\ 
			(7680x4320)}}&\multirow{2}{*}{\it SubwayTree}& -3.55 (-17.17) & -1.15 (-9.12) & -2.19 (-5.85)	& -0.90 (-3.39)	 \\
			&\multirow{2}{*}{\it TiergartenParkway}& -0.88 (-7.90) & -0.93 (-5.46) & -1.64 (-3.26) & -0.80 (-1.99) \\
			\midrule[0.2mm]
			\multirow{6}{*}{\shortstack[l]{ClassA1 \\ (3840x2160)}}&\multirow{2}{*}{\it Campfire}& -1.19 (-9.51) & -0.78 (-6.16) & -1.30 (-2.92)	& -0.58 (-1.53)	 \\
			&\multirow{2}{*}{\it FoodMarket4}& -1.97 (-15.97) & -1.55 (-9.66) & -2.19 (-5.13) & -0.97 (-2.67) \\
			&\multirow{2}{*}{\it Tango2}& -1.47 (-8.41) & -1.40 (-5.61) & -3.91 (-6.04) & -1.19 (-3.23) \\
			\midrule[0.2mm]
			\multirow{6}{*}{\shortstack[l]{ClassA2 \\ (3840x2160)}}&\multirow{2}{*}{\it CatRobot1}&-2.47 (-16.75)&-2.53 (-15.87)& -2.41 (-5.62)& -1.47 (-3.87)	 \\
			&\multirow{2}{*}{\it DaylightRoad2}& -1.37 (-10.28) & -1.31 (-7.30) & -1.89 (-4.77) & -0.94 (-2.44) \\
			&\multirow{2}{*}{\it ParkRunning3}& -1.44 (-12.45) & -1.34 (-10.15) & -1.46 (-3.18) & -0.84 (-2.34) \\
			\midrule[0.2mm]
			\multirow{10}{*}{\shortstack[l]{ClassB \\ (1920x1080)}}&\multirow{2}{*}{\it BasketballDrive}&-5.44 (-53.94)&-4.22 (-40.42)& -1.92 (-4.29)& -1.68 (-3.59)	 \\
			&\multirow{2}{*}{\it BQTerrace}& -7.99 (-55.00) & -5.43 (-34.09) & -2.64 (-4.97) & -1.58 (-2.82) \\
			&\multirow{2}{*}{\it Cactus}& -3.01 (-33.17) & -2.35 (-22.93) & -2.16 (-4.55) & -1.37 (-3.24) \\
			&\multirow{2}{*}{\it MarketPlace}& -3.52 (-28.84) & -1.63 (-18.10) & -3.10 (-5.25) & -0.78 (-2.64) \\
			&\multirow{2}{*}{\it RitualDance}& -2.54 (-17.50) & -2.55 (-12.82) & -3.61 (-7.10) & -1.66 (-5.08) \\
			\midrule[0.3mm]
			&\multirow{2}{*}{\textbf{Average}}&\textbf{-2.83 (-22.07)}&\textbf{-2.09 (-15.21)}&\textbf{-2.34 (-4.84)}&\textbf{-1.14 (-2.99)} \\
			\midrule[0.3mm]
		\end{tabularx}
		}
	\end{center}
	\vspace{-2.5em}
\end{table}

%0.6033 sec for an HD image

\section{Conclusion}
\label{sec:conclusion}

In this work, we presented a multitask learning-based approach that performs both super-resolution and quality enhancement of \gls*{VVC} intra-coded frames. We used a multi-QPs training strategy based on fine-tuning and prior information. We demonstrated that our method allows a significant reduction of parameters, while maintaining a good quality of reconstruction compared to specialized solutions. We also showed that our approach offers quality enhancements compared to single-task models when the same total number of parameters is considered. As future work, we plan to include the temporal aspect into our model to ensure temporal consistency and enable inter-coded frame processing. In addition, the integration of our model in the coding loop will also be investigated in order to perform both quality enhancement and super-resolution of the base layer directly into a scalable codec using a single shared network.

\bibliographystyle{IEEEtran}
\bibliography{IEEEexample}

\end{document}